\documentclass[conference]{IEEEtran}
\IEEEoverridecommandlockouts

\usepackage{cite}
\usepackage{amsmath,amssymb,amsfonts}
\usepackage{algorithmic}
\usepackage{algorithm}
\usepackage{graphicx}
\usepackage{textcomp}
\usepackage{xcolor}
\usepackage{url}
\usepackage{tikz}
\def\BibTeX{{\rm B\kern-.05em{\sc i\kern-.025em b}\kern-.08em
    T\kern-.1667em\lower.7ex\hbox{E}\kern-.125emX}}
\DeclareMathOperator*{\argmax}{arg\,max}

\newcommand\copyrighttext{%
  \footnotesize \copyright 2018 IEEE. Personal use of this material is permitted. Permission from IEEE must be
obtained for all other uses, in any current or future media, including
reprinting/republishing this material for advertising or promotional purposes, creating new
collective works, for resale or redistribution to servers or lists, or reuse of any copyrighted
component of this work in other works.}
\newcommand\copyrightnoticeb{%
\begin{tikzpicture}[remember picture,overlay]
\node[anchor=south,yshift=10pt] at (current page.south) {\fbox{\parbox{\dimexpr\textwidth-\fboxsep-\fboxrule\relax}{\copyrighttext}}};
\end{tikzpicture}%
}

\newcommand{\R}{\mathbb{R}}                      

\begin{document}

\title{Too many secants: a hierarchical approach to secant-based dimensionality reduction on large data sets\\
\thanks{This paper is based on research partially supported by the National Science Foundation under Grants No. DMS-1513633, 
DMS-1322508,  as well as  DARPA awards  N66001-17-2-4020 and D17AP00004.
} 
}

\author{\IEEEauthorblockN{Henry Kvinge, Elin Farnell, Michael Kirby, and Chris Peterson}
\IEEEauthorblockA{Department of Mathematics\\
Colorado State University\\Fort Collins, CO 80523-1874}}

\maketitle

\copyrightnoticeb

\begin{abstract}
A fundamental question in many data analysis settings is the problem of discerning the ``natural'' dimension of a data set. That is, when a data set is drawn from a manifold (possibly with noise), a meaningful aspect of the data is the dimension of that manifold. Various approaches exist for estimating this dimension, such as the method of Secant-Avoidance Projection (SAP). Intuitively, the SAP algorithm seeks to determine a projection which best preserves the lengths of all secants between points in a data set; by applying the algorithm to find the best projections to vector spaces of various dimensions, one may infer the dimension of the manifold of origination. That is, one may learn the dimension at which it is possible to construct a diffeomorphic copy of the data in a lower-dimensional Euclidean space. Using Whitney's embedding theorem, we can relate this information to the natural dimension of the data. A drawback of the SAP algorithm is that a data set with $T$ points has $O(T^2)$ secants, making the computation and storage of all secants infeasible for very large data sets. In this paper, we propose a novel algorithm that generalizes the SAP algorithm with an emphasis on addressing this issue. That is, we propose a hierarchical secant-based dimensionality-reduction method, which can be employed for data sets where explicitly calculating all secants is not feasible.
\end{abstract}


\section{Introduction}
Determining the dimension of a data set is a basic first step toward a meaningful understanding of the data as well as a foundational part of any related data analysis. This is especially true for high-dimensional data sets where calculating the ``intrinsic'' dimension can point toward huge efficiency gains via dimensionality reduction.  The determination of dimension is a central question in the area of geometric data analysis~\cite{kirby_wiley2}  and the related field of manifold learning~\cite{Tenenbaum2319,RoSa00}.   This paper presents a subspace-secant algorithm for computing projections of data which preserves
both the topological and Hausdorff dimension of a data set.  A byproduct of the computation is an
estimate of the dimension and an
estimate of the smoothness, and hence stability, 
of the nonlinear mapping that reconstructs the data.

In \cite{KFKP18} the authors described the SAP (Secant-Avoidance Projection) algorithm. This algorithm is based in part on the commonsense notion that a good dimensionality-reduction algorithm should strive to be distance preserving. That is, if two data points are initially far apart, then when we map them into a reduced space they should remain far apart. This can be rephrased to say that a dimensionality-reduction algorithm should preserve the lengths of the secants between points in the data set. The SAP algorithm takes as input the secant set $S$ of a data set in $\R^n$ and an integer $0 < k < n$ and produces a matrix $P: \R^n \rightarrow \R^k$ corresponding to a projection $PP^T$ maximizing the value $\min_{s \in S}||PP^Ts||_{2}$. Furthermore, by studying the quality of projections produced by SAP over a range of projection dimensions we can make a reasonable estimate of the dimension of our data.  

One limitation of the SAP algorithm is that it does not scale well to large data sets because the number of secants corresponding to a data set of size $T$ is $T(T-1)/2$. The motivation for this paper is to propose a generalization of the SAP algorithm, which we call the HSAP (Hierarchical Secant-Avoidance Projection) algorithm, which addresses this limitation. The underlying idea of the HSAP algorithm is to use the hierarchy of structure present in a data set in order to reduce the number of secants required to obtain a good secant preserving projection. In this case this means first clustering the data set and then either using a linear approximation of each cluster or a sample of secants from each cluster to capture the secant structure at the local level. At the same time we also sample a fairly small number of secants between clusters to capture the spatial relations between clusters. Roughly then, the HSAP algorithm generates a projection that neither maps the points of a single cluster onto each other nor maps one cluster onto another. 
 
The outline of the paper is as follows: in Section \ref{section-background} we review some of the mathematical framework that underlies this paper including the geometry of Grassmann manifolds and dimension estimation. In Section \ref{section-algorithm} we describe two versions of the HSAP algorithm and remark on various aspects of it. Finally in Section \ref{section-examples} we describe the result of running the HSAP algorithm on both a small-dimensional synthetic data set and also a hyperspectral data set. 

\section{Background} \label{section-background}
Finding compact representations for complicated objects, such as data clouds or high-dimensional arrays, has been an indispensable tool for knowledge discovery within massive data sets. For instance, if a cloud of points cluster along a $k$-dimensional linear space within a vector space, then for many purposes it is natural to represent the cloud with its linear approximation. If a more refined representation is desired then one approach is to consider the linear space that captures a pre-specified percentage of the energy in the cluster together with a sparse sampling of the points within the cluster which captures some essence of the distribution of the data. The {\sl Grassmannian} $Gr(k,n)$ is a manifold whose points parametrize the $k$-dimensional subspaces of a fixed $n$-dimensional vector space. Using this manifold, the cluster can be represented by a point on $Gr(k,n)$ together with the sparse sampling representing the distribution. Suppose now that one would like to condense the information within a very large data cloud residing in a high-dimensional vector space. In many applications, such a data cloud can be hierarchically partitioned into a collection of smaller clusters with each representing some feature of interest. By compactly representing the points in each cluster and the relationships between the clusters, one can hope to better understand the data cloud as a whole. In the sections that follow, we utilize Grassmann manifolds as organizing structures to condense and capture much of the information in a very large data cloud and use this compact representation of the data cloud to drive algorithms towards locally optimal, dimensionality-reducing, structure-preserving projections. 

An important feature of Grassmannians, in the context of knowledge discovery in data, is that they can be given the structure of a differentiable manifold. One would like to determine the proximity of various points on a Grassmannian and this is typically carried out by first determining principal angles between the corresponding vector spaces.  This derives from the fact that every orthogonally invariant metric on a Grassmann manifold can be described in terms of principal angles.  Furthermore, principal angles between vector spaces are readily computable through a singular value computation. In order to understand this statement, we first describe principal angles in the context of an optimization procedure.

Consider the subspaces $U$ and $V$ of a vector space $\mathbb R^n$ and let $q= \min\left\{\dim U,\dim V \right\}$.
The principal angles between $U$ and $V$ are the angles $\theta_1, \theta_2, \dots \theta_q \in [0,\frac{\pi}{2}]$ between pairs of principal vectors $\{ u_k, v_k \}$
with $u_1, \dots, u_q$ a distinguished orthonormal set of vector in $U$ and $v_1, \dots, v_q$ a distinguished set of orthonormal vectors in $V$. These vectors are obtained recursively, for each $1\leq k \leq q$, by defining
\[
\cos \theta_k = \underset{u \in U, v \in V}{\max} u^T v = u_k^T v_k
\]
subject to 
\begin{itemize}
\item $||u||_2 = ||v||_2 = 1$
\item $u^T u_i = 0$ and $v^T v_i = 0$ for $i = 1,2, \ldots, k-1.$
\end{itemize}
The key point is that any orthogonally invariant measure of similarity between $U$ and $V$ can be determined as a function of the principal angles.  

The principal angles and principal vectors between $U$ and $V$ can be determined from orthonormal bases for $U$ and $V$ as follows.
Suppose $A$ (respectively $B$) are matrices whose columns form orthonormal bases for $U$ (respectively $V$). From the singular value decomposition we have a factorization $A^TB=Y\Sigma Z^T$. If $y_i$ (respectively $z_i$) denotes the $i^{th}$ column of $Y$ (respectively $Z$) then the $i^{th}$ singular vector pair can be computed as $u_i=Ay_i$ and $v_i=Bz_i$. Furthermore, the singular values of $A^TB$ are equal to $\cos \theta_1, \cos \theta_2, \dots, \cos \theta_q,$ where the sequence is assumed to be monotonically decreasing. See \cite{bjorck1973numerical}.

\subsection{Dimension Estimation}
The estimation of dimension from data has been addressed by numerous authors
including, e.g.,  \cite{broomhead_jones_king87,AnHuKi02,kirby_hundley1999,costa2006determining,fukunaga1971algorithm,BK05,camastra2003data,cunningham2015linear,wang2015survey}.

There is a useful theoretical result for characterizing
dimension-preserving transformations.  It revolves around the definition
of a  {\it bi-Lipschitz} function.
A function $f(x)$ is said to be {\it bi-Lipschitz} on $X$ if for all $x,y \in X$ it
holds that
$$
a \| x - y \|_{\ell_2} \le \| f(x) - f(y) \|_{\ell_2} \le  b \| x - y \|_{\ell_2}.
$$
The constant $a$ restricts pairs of points from collapsing on top of each other while $b$ restricts pairs of points from blowing apart.  In the context of projection, as we consider in the algorithm of the next section, we can restrict to the case $b=1$. A key feature of bi-Lipshitz functions is:
$$
{\rm if}\ f:X \rightarrow Z \ {\textrm{ is bi-Lipschitz, then}} \ 
\dim(X) =  \dim(Z)
$$
where the dimension can be taken
as the topological dimension, or the Hausdorff dimension; 
see~\cite{Fal03} for details.
Thus we see a link between dimension preservation and projections that
avoid collapsing secants.  Projection-based algorithms
that maximally avoid decreasing the length of secants are, in some sense, optimally dimension preserving 
and form the theoretical motivation for the algorithm presented here.
An additional  argument for this approach, based on invoking Whitney's easy embedding theorem, is made in \cite{kirby_1998a}.

\section{The algorithm} \label{section-algorithm}

We begin this section by noting two different methods of representing the secant set of a cluster: either by a linear approximation (Section \ref{subsection-approx-by-linear-subspaces}) or by a sampling of secants (Secant \ref{subsection-approx-by-subsets}). We then go on to describe the HSAP algorithm proper (Section \ref{general-algorithm}).

\subsection{Approximation by linear subspaces} 
\label{subsection-approx-by-linear-subspaces}

Let $D$ be a set of points in $\R^n$. Using a clustering algorithm one can partition $D$ into $N$ disjoint subsets $D_1, D_2, \dots, D_N$ (see Section \ref{subsection-clustering-subspace} for a discussion of clustering methods). To each $D_j$ for $1 \leq j \leq N$ we construct a $k_j$-dimensional linear approximation. That is, to each of these $N$ linear approximations, we associate an $n \times k_j$ matrix $V_j = [v_1^{(j)},v_2^{(j)},\dots,v_{k_j}^{(j)}]$ whose columns form an orthonormal basis for the linear approximation subspace. In particular each $V_j$ also approximates the secant set for points in $D_j$. We suggest mean-centering each cluster $D_j$ and using Principal Component Analysis (\hspace{1sp}\cite{H33,J86}) to determine a good $V_j,$ and we note that this approach is appropriate precisely when the cluster and consequently the secant set are well approximated as linear spaces.

In Algorithm \ref{algo-}, we present this version of the HSAP algorithm, which is the more involved of the two. The modifications to Algorithm \ref{algo-} required for the version described below in Section \ref{subsection-approx-by-subsets} should be clear from the context.

\subsection{Approximation by sampled secants}\label{subsection-approx-by-subsets}

In the case where clusters are expected to be highly non-linear in structure, it makes sense to take an approach which takes this into account. Therefore in the second version of our algorithm, instead of approximating our clusters $D_1, \dots, D_N$ by linear spaces, we instead approximate the secant set of each of these clusters by a subset of its secant set. For cluster $D_j$ let $S_j$ be the corresponding secant subset, and include each $S_j$ in the set of secants $\tilde{S}$ defined below.





\subsection{The HSAP Algorithm} \label{general-algorithm}

In order to encode the relations between different clusters, we also sample a small selection of points $A_j$ from $D_j$ for $1 \leq j \leq N$. There are different strategies for doing this, e.g. one might collect a random sample from $D_j$ or one might select extremal points of $D_j$. We calculate all secants between points in $A_i$ and points in $A_j$ for $1 \leq i < j \leq N$. Define $\tilde{S}$ to be the subset of the full secant set that consists of all secants between points in each pair $(A_i,A_j),i\neq j$. 
Finally, we choose an initial $k$-dimensional projection with a corresponding $n \times k$ matrix $P^{(0)}$ whose orthonormal columns span the projection subspace. We propose an initialization $P^{(0)}$ defined as the first $k$ columns of $Y,$ where the data matrix has been decomposed via the singular value decomposition as $Y \Sigma Z^T.$

To obtain a matrix $P^{(i+1)}$ such that the projection $P^{(i+1)(P^{(i+1)})^T}$ better preserves secants or their approximations, at the $i$th iteration we shift our current matrix $P^{(i)}$ toward the secant (or corresponding approximation) which is currently the worst preserved by $P^{(i)}$. We call this vector the ``shortest representative vector'' and denote it by $w_{i}$ (note that it can come from either a linear approximation of a secant set or a genuine secant sampled  between clusters). We will also use the projection (which we denote by $w^{(p)}_i$) of this shortest representative vector onto the subspace corresponding to $P^{(i)}$. In order to find which representative vector is worst preserved, 
\begin{itemize}
\item calculate the singular values $\sigma_1^{(j)}, \dots , \sigma_{k_j}^{(j)}$ of $(P^{(i)})^T V_j$ for each $1 \leq j \leq N$,
\item calculate the length $||(P^{(i)})^T s||_{\ell_2}$ for each $s \in \tilde{S}$.
\end{itemize}
Note that $\sigma_1^{(j)}, \dots , \sigma_{k_j}^{(j)}$ correspond to the cosine function applied to the principal angles between the subspaces corresponding to $P^{(i)}$ and $V_j$. This is a natural higher-dimensional generalization of the process of measuring the length of a unit vector projected onto a subspace (this is one sense in which the HSAP algorithm is a generalization of the SAP algorithm). 
We now calculate the minimum element of the set $R$ defined below and define the shortest representative vector $w_i$ and its projection accordingly: $$\displaystyle R=\Big(\bigcup_{j = 1}^N\{\sigma_1^{(j)}, \dots , \sigma_{k_j}^{(j)}\}\Big) \bigcup \Big(\bigcup_{s \in \tilde{S}}\{||(P^{(i)})^T s||_{\ell_2} \}\Big) .$$

\begin{itemize}
\item \textbf{Case 1:} If the smallest element is $\sigma_{k_j}^{(j)},$ let $y_{k_j}^{(j)}$ and $z_{k_j}^{(j)}$ be the corresponding left and right singular vectors, respectively, in the SVD of $(P^{(i)})^TV_j$. Then $w^{(p)}_i = P^{(i)}y_{k_j}^{(j)}$ and $w_i = V_jz_{k_j}^{(j)}$. Note that we assume that the singular values of $(P^{(i)})^T V_j$ are ordered from largest to smallest as in the standard singular value decomposition. In this case we only have to calculate the last singular value for the comparison step. 
\item \textbf{Case 2:} If $||(P^{(i)})^Ts||_{\ell_2}$ is the smallest element, then $w_i = s$ and $w^{(p)}_i = P^{(j)}(P^{(j)})^Ts$.
\end{itemize}

Finally, we construct $P^{(i+1)}$ from $P^{(i)}$ by first finding the column $P^{(i)}_t$ of $P^{(i)}$ such that $|(P^{(i)}_q)^Ts^*|$ is maximized over all columns $P_q^{(i)}$. Assume that $\max_{P_q^{(i)}}|(P^{(i)}_q)^Ts^*| > 0$ (we will treat the special case where $||(P^{(i)})^Tw_i|| = 0$ below). We then remove the $t$-th column of $P^{(i)}$, shift all columns with index strictly less than $t$ forward and add $w^{(p)}_i$ as the first column. We run the Gram-Schmidt algorithm on this new matrix to obtain a matrix $\hat{P}^{(i)}$ whose columns are orthonormal. Note that by construction $P^{(i)}$ and $\hat{P}^{(i)}$ project to the same subspace. $P^{(i+1)}$ is then the matrix obtained by replacing the first column $\hat{P}^{(1)}_1$ by the normalization of $(1-\alpha)\hat{P}^{(1)}_1 + \alpha(w_i - \hat{P}^{(1)}_1)$ where $\alpha \in [0,1]$ is small.

In the case where $||(P^{(i)})^Tw_i|| = 0,$ we replace the first column $P^{(i)}_1$ of $P^{(i)}$ with $(1-\alpha)P^{(i)}_1 + \alpha w_i$ and run the Gram-Schmidt algorithm on the resulting matrix. The result is $P^{(i+1)}$. 

\begin{algorithm} 
\caption{\label{algo-} Hierarchical Secant-Avoidance Projection}
\begin{algorithmic}[1]
\STATE \textbf{inputs} Given a data set $D$.\\ Initialize parameters: ambient dimension $n$, number of clusters $N$, max number of steps (Iterations) or alternative stopping criterion, and shift parameter $\alpha$.\\
Use a clustering algorithm to find $N$ clusters $D_1, D_2,\ldots, D_N$ that partition $D.$\\
Define matrices $V_j$ for $1\leq j\leq N,$ whose orthonormal columns $\big\{v_i^{(j)}\big\}_{i=1}^{k_j}$ form bases for the linear approximations to clusters $D_1, D_2, \ldots, D_N$. Choose small subsets $A_1, A_2, \dots, A_N$ of $D_1, D_2, \dots, D_N,$ respectively. Choose an initial matrix $P^{(0)}$ in $\mathbb{R}^{n\times k}.$ Define $\tilde{S}$ to be the set of secants between all points in each pair of sets $(A_i,A_j)$ with $i\neq j.$
\STATE Calculate all secants between points in $A_i$ and $A_j$ for all $1 \leq i < j \leq N$. 
\FOR{$i \leq $ Iterations} 
\STATE \label{algo-singular-value-decomp} Calculate the singular values $\sigma_1^{(j)}, \dots , \sigma_{k_j}^{(j)}$ for $(P^{(i)})^T V_j$ for each $1 \leq j \leq N$. 
\STATE \label{algo-length-calc} Calculate the length of $(P^{(i)})^T s$ for each $s \in \tilde{S}$. 
\STATE \label{choose-smallest} Choose the smallest value among $\sigma_1^{(j)}, \dots , \sigma_{k_j}^{(j)}$ for $1 \leq j \leq N$ and $||(P^{(i)})^T s||_{\ell_2}$ for all $s \in \tilde{S}$ in steps \ref{algo-singular-value-decomp}-\ref{algo-length-calc} above.
\STATE Calculate $w_i$ and $w^{(p)}_i$ as in Section \ref{general-algorithm}.
\IF{$w_i^{(p)} = 0$}
\STATE Replace the first column $P^{(i)}_1$ of $P^{(i)}$ by $(1-\alpha)P^{(i)}_1 + \alpha w_i$, run the Gram-Schmidt algorithm on the resulting matrix and set the result equal to $P^{(i+1)}$.
\ELSE
\STATE Set $t = \argmax_{1 \leq q \leq k} |(P^{(i)}_q)^T w_i|$. 
\STATE Apply the modified Gram-Schmidt algorithm to $w^{(p)}_i, P^{(i)}_1, \dots, P^{(i)}_{t-1}, P^{(i)}_{t+1},\dots, P^{(i)}_k$ to obtain a new orthonormal projection $\hat{P}^{(i)}$.
\STATE Replace the first column of $\hat{P}^{(i)}$ with the normalization $(1-\alpha) \hat{P}^{(i)}_1 + \alpha(w-\hat{P}^{(i)}_1).$
\STATE $i+1 \leftarrow i$
\ENDIF
\ENDFOR
\RETURN 
\end{algorithmic}
\end{algorithm}

\subsection{Remarks on the HSAP algorithm} \label{subsection-clustering-subspace}

There are a number of decisions which must be made when applying the HSAP algorithm to a data set. 

A choice of parameter $\alpha$ must be made, which controls the extent to which the projection shifts at each step. In practice we have found that when the algorithm is run using $\alpha$ values between $0.01$ and $0.05$, convergence occurs reasonably quickly but still reliably. 

The HSAP algorithm takes as input clusters $D_1,D_2,\ldots,D_N$ for a data set $D.$ Thus a fundamental step in the application of the HSAP algorithm is clustering data. In certain examples, one may have knowledge of clusters withing the data a priori; in general, we must expect that it will be necessary to apply an algorithm that attempts to cluster the data in an optimal way. In light of the fact that the HSAP algorithm is designed to succeed in a scenario in which the data set $D$ is very large, we note that there are many relevant clustering algorithms that are designed for the big data setting. For example, in \cite{HB06}, the authors propose extensions to fuzzy and probabilistic clustering for big data settings. Many authors have suggested implementations of $k$-means clustering that utilize graphics processors for efficiency, e.g. \cite{CTZ06,WZH09,LZCL13,HLDZH09,SDT08,FRCC08}. See \cite{SAWH14,FATAKZFB14} for reviews of clustering methods for big data. We choose to use a $k$-means algorithm in the work we present in this paper; we leave it to the reader to select an appropriate clustering algorithm for the particular data setting of interest.

There are several options for the initial matrix $P^{(0)}.$ An efficient choice would be to define $P^{(0)}$ to be a random matrix with orthonormal columns. In practice, this seems to suffice, though convergence often requires more iterations when compared with our proposed initialization (the truncated PCA basis for the column space of the data matrix).

In Step \ref{choose-smallest} of the HSAP algorithm, it is necessary to compare the smallest singular values for each $V_i$ against other scalars. We note that it is possible to compute only the smallest singular value as a means of added computational efficiency. There are several articles containing algorithms to do this (e.g. \cite{SS03,H01,LC15}) and there are implementations in popular programming languages, such as MATLAB\textsuperscript{\textregistered}
 \cite{Matlab}, which relies on \cite{BR05} and \cite{L98}.
 
The HSAP algorithm is a polynomial time algorithm. The complexity is dominated by the computation of the projected cluster bases and of the projection of the secants. We note that, while an SVD is often an expensive computation, in the HSAP algorithm it does not dominate because the relevant matrices are comparatively small: the computation of the SVD in Step \ref{algo-singular-value-decomp} is $O(\min{\left\{k^2k_j,kk_j^2\right\}}).$ Meanwhile, the computation of the products $(P^{(i)})^TV_j$ for all $j=1,\ldots,N$ is $O(knk_jN)\subseteq O(n^3N).$ The computation of the norm of the shortest projected secant in Step \ref{choose-smallest} requires computing products $(P^{(i)})^Ts$ for all $s\in \tilde{S};$ these products are $O(kn)\subseteq O(n^2).$ Since $|\tilde{S}|$ is $O(N^2(\max{|A_i|})^2),$ the computation of all products is $O(n^2N^2(\max{|A_i|})^2).$ Thus, the HSAP algorithm is $O(\max{\left\{n^3N,n^2N^2(\max{|A_i|})^2\right\}}).$ In practice, we expect that $\max(|A_i|)$ and $N$ will usually be substantially smaller than $n$. 

\section{Examples} \label{section-examples}

\subsection{A synthetic example}

\begin{figure}
\includegraphics[width=8cm, height=6cm]{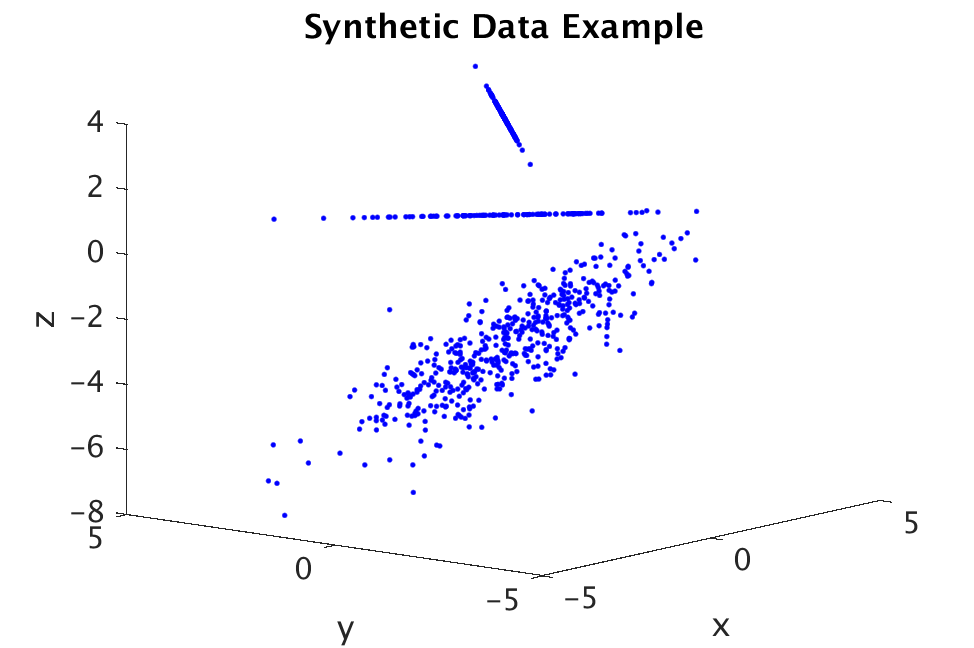}
\caption{\label{figure-3D-dataset} The synthetic data set $D_{\text{syn}}$ in $\R^3$. We construct it to have three natural clusters; the HSAP algorithm should prevent the clusters from collapsing onto each other while simultaneously seeking to preserve the data within each cluster in the projection.}
\end{figure}

We construct a data set $D_{\text{syn}}$ in $\R^3$ consisting of the union of $100$ points sampled from two different lines and $500$ points sampled from a plane. Specifically we sample points from
\begin{align*}
&f_1(t) = \Big\langle t, -t, 1 \Big\rangle \\
&f_2(t) = \Big\langle t, t, 4 \Big\rangle \\
&f_3(t,s) = \Big\langle \frac{t}{2} - s, s, t - s - 3 \Big\rangle
\end{align*}
(see Figure \ref{figure-3D-dataset}). Note that in this case $D_{\text{syn}}$ is naturally clustered as subspaces of varying dimensions (this is an artificial situation but serves well as a first illustration of the algorithm). 

We ran $80$ iterations of the linear approximation version of on the HSAP algorithm on $D_{\text{syn}}$ to obtain a matrix $P^{(80)}$ that maps from $\R^3$ into $\R^2$. Each sample of points from within a cluster was chosen randomly and had a size of $20$ (i.e. $|A_j| = 20$ for $j = 1,2,3$). We also set $\alpha = 0.01$.

\begin{figure}
\includegraphics[width=8cm, height=6cm]{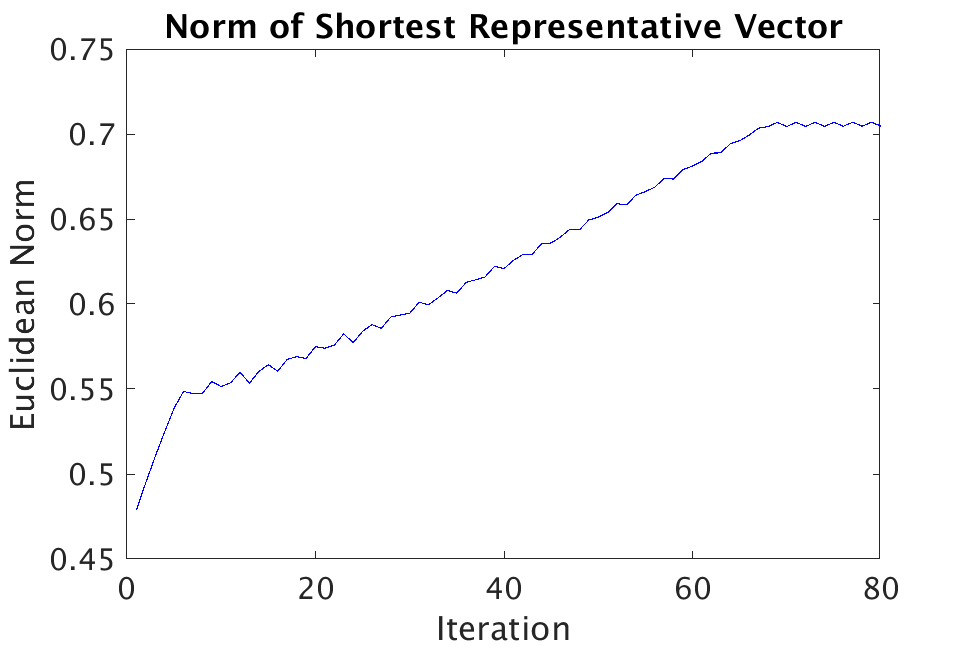}
\caption{\label{figure-convergence} A plot of the convergence of the linear approximation version of the HSAP algorithm run on the $D_{\text{syn}}$ data set. The iteration is given on the $x$-axis, while the $y$-axis gives the smallest singular value of $(P^{(i)})^TV_j$ or the smallest representative vector length.}
\end{figure}

In Figure \ref{figure-convergence} we plot the norm of the projection of the shortest representative vector as a function of iteration. As can be seen, the projection improves fairly quickly over the course of approximately 70 iterations but then stalls in what is probably a local minimum.

\begin{figure}
\includegraphics[width=8cm, height=6cm]{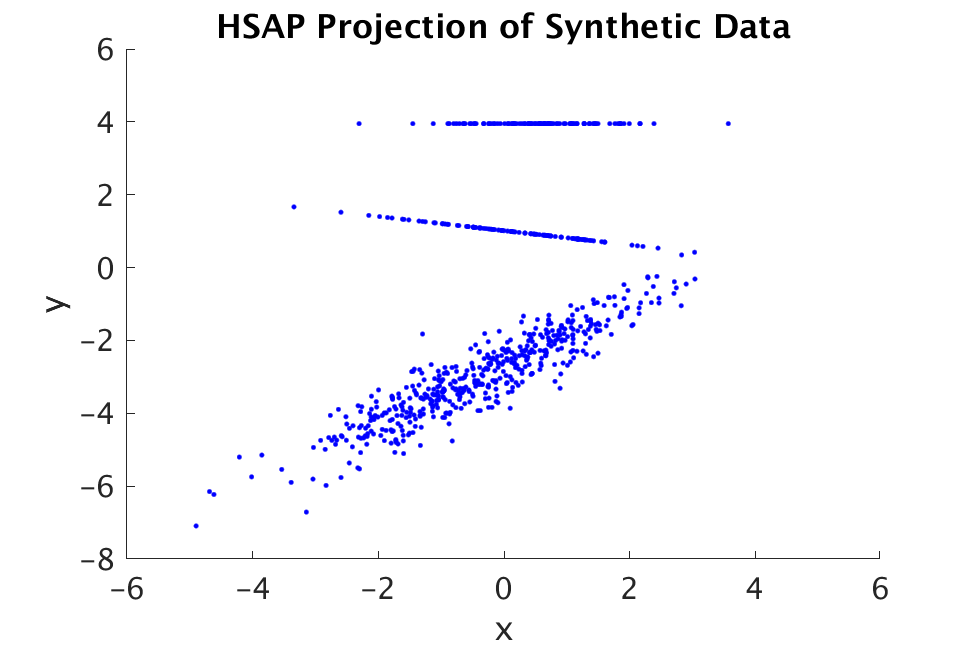}
\caption{\label{figure-2d-proj} The projection of the data set $D_{\text{syn}}$ using the output $P^{(80)}$ of an application of the linear approximation version of the HSAP algorithm.}
\end{figure}

We see the results of the HSAP projection of $D_{\text{syn}}$ into $\mathbb{R}^2$ in Figure \ref{figure-2d-proj}. Note that the algorithm has successfully projected the natural clusters in the data into distinct locations in $\mathbb{R}^2$ while also preserving the within-cluster spacing to a reasonable extent.

\subsection{The Indian Pines data set}

As a real-world example we apply the HSAP algorithm to the Indian Pines hyperspectral data set (some bands covering the region of water absorption are removed) \cite{IP}. The data cube is $145 \times 145 \times 200;$ that is, there are $200$ bands, each with spatial resolution of $145 \times 145$. We define a data set $D\subset\mathbb{R}^{200}$ to be the collection of vectors of spectral information taken across all pixel locations. We then have $|D|=21,025.$ There are consequently over 221 million secants for this data set. What is more, one would expect that the Indian Pines data set has a naturally clustered structure, where clusters correspond to materials with different absorbency in the scene. These two qualities make the Indian Pines data an ideal candidate for the HSAP algorithm.

\begin{figure}
\includegraphics[width=10cm, height=7cm]{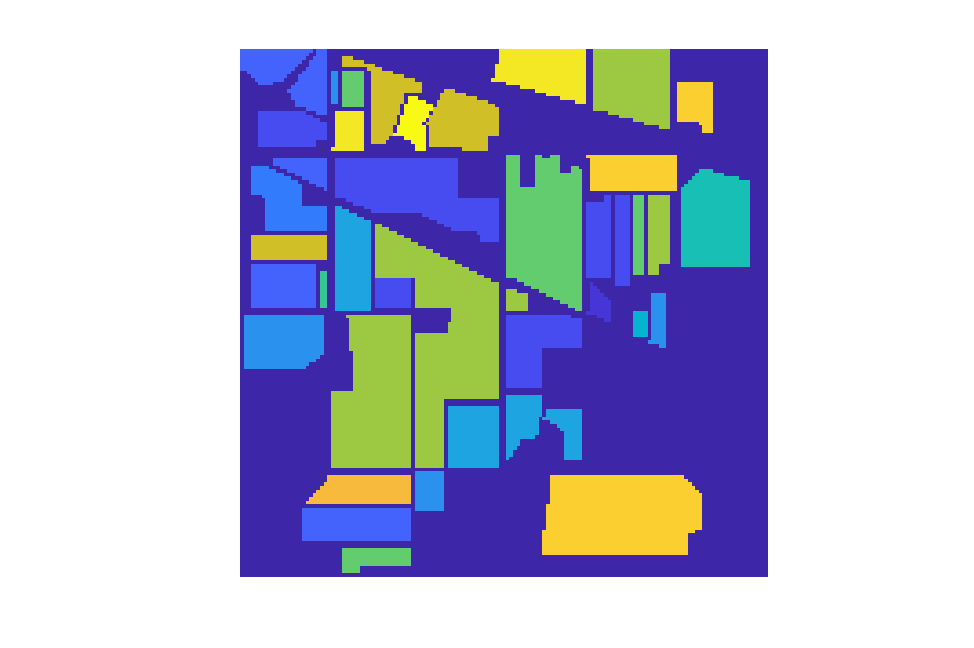}
\caption{\label{figure-indianpines-gt} The ground truth labels for the Indian Pines hyperspectral data set. There are 16 labeled categories, including, e.g. alfalfa, corn, oats, woods, and stone-steel-towers. There is also a 17th category of unclassified pixels.}
\end{figure}

The manually-labeled ground truth available for the Indian Pines data set is displayed in Figure \ref{figure-indianpines-gt}. There are 16 labeled categories (e.g. alfalfa, oats, woods, and buildings-grass-trees-drives) and a 17th category of unclassified pixels. 

\begin{figure}
\includegraphics[width=10cm, height=7cm]{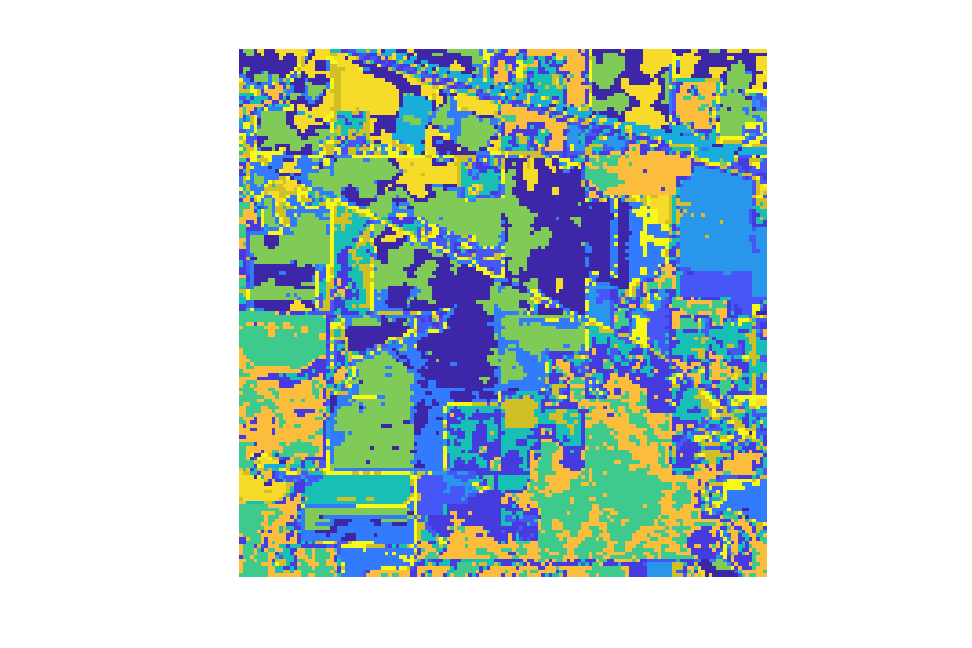}
\caption{\label{figure-indianpines-clusters} The 13 clusters determined by an application of the $k$-means algorithm to the Indian Pines hyperspectral data using cosine distance. We note that this clustering visually appears to have captured some of the relevant structure in the data set as shown in Figure~\ref{figure-indianpines-gt}.}
\end{figure}

\begin{figure}
\includegraphics[width=9cm, height=8cm]{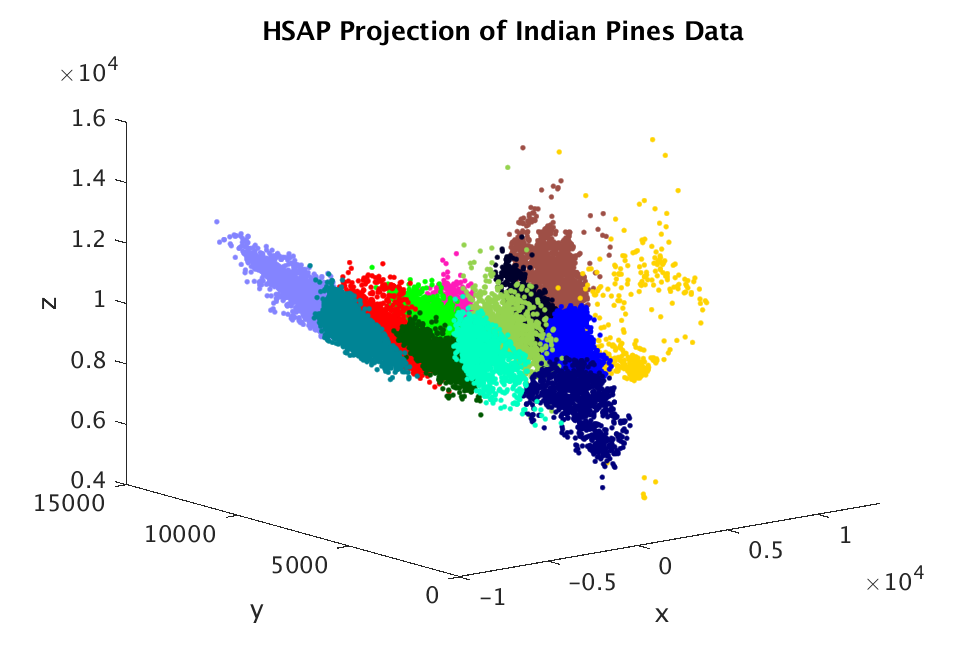}
\caption{\label{figure-3d-proj} The projection into $\mathbb{R}^3$ defined by the HSAP algorithm when applied to the 13 data clusters of Indian Pines hyperspectral data shown in Figure \ref{figure-indianpines-clusters}. Note that the HSAP algorithm appears to have successfully preserved the clusters in the provided projection.}
\end{figure}

In Figure, \ref{figure-3d-proj}, we see the result of projecting the Indian Pines data with the result of the HSAP algorithm. In this example, we project into $\mathbb{R}^3,$ and we choose to approximate the clusters with linear spaces (see Section \ref{subsection-approx-by-linear-subspaces}). In order to define the input clusters $D_1,D_2,\ldots,D_N$, we apply the $k$-means clustering algorithm with cosine distance to get approximations of the naturally occurring clusters in the data. Note that cosine distance $d$ is defined to be $d(u,v)=1-\cos(\theta),$ where $\theta$ is the angle between the two input vectors $u$ and $v.$ See Figure \ref{figure-indianpines-clusters} for the visualization of the result of this application of $k$-means. In Figure \ref{figure-3d-proj}, the individual clusters are assigned different display colors. Note that the projection appears to do a very good job of preventing these clusters from being collapsed together while also maintaining some spread among the points within clusters.  


\section{Conclusion}
In this paper, we proposed a generalization of the Secant-Avoidance Projection (SAP) algorithm that is particularly relevant to very large data sets. The Hierarchical Secant-Avoidance Projection (HSAP) algorithm uses a structured approach to selecting appropriate subsets and approximations to the full secant set in order to guide an iterative algorithm. The algorithm returns a projection, which seeks to best preserve the secant set associated to a data set. This projection is useful for both dimensionality reduction and for approximating the dimension of the manifold from which the data was drawn (see \cite{KFKP18} for more on this), when such a setting exists. The usefulness and relevance of the algorithm was demonstrated in a synthetic example and in an application to the Indian Pines hyperspectral data set. 

Further research can be completed in the following areas.
\begin{itemize}
\item Other methods of selecting subsets of secants should be considered.
\item We proposed one of many hierarchical structures. There are opportunities to explore alternatives.
\end{itemize}


\bibliographystyle{IEEEtran}
\bibliography{refs,complete5,KirbyMay2018,books}

\begin{thebibliography}{10}
\providecommand{\url}[1]{#1}
\csname url@samestyle\endcsname
\providecommand{\newblock}{\relax}
\providecommand{\bibinfo}[2]{#2}
\providecommand{\BIBentrySTDinterwordspacing}{\spaceskip=0pt\relax}
\providecommand{\BIBentryALTinterwordstretchfactor}{4}
\providecommand{\BIBentryALTinterwordspacing}{\spaceskip=\fontdimen2\font plus
\BIBentryALTinterwordstretchfactor\fontdimen3\font minus
  \fontdimen4\font\relax}
\providecommand{\BIBforeignlanguage}[2]{{%
\expandafter\ifx\csname l@#1\endcsname\relax
\typeout{** WARNING: IEEEtran.bst: No hyphenation pattern has been}%
\typeout{** loaded for the language `#1'. Using the pattern for}%
\typeout{** the default language instead.}%
\else
\language=\csname l@#1\endcsname
\fi
#2}}
\providecommand{\BIBdecl}{\relax}
\BIBdecl

\bibitem{kirby_wiley2}
M.~Kirby, \emph{Geometric Data Analysis: An Empirical Approach to
  Dimensionality Reduction and the Study of Patterns}.\hskip 1em plus 0.5em
  minus 0.4em\relax Wiley, 2001.

\bibitem{Tenenbaum2319}
\BIBentryALTinterwordspacing
J.~B. Tenenbaum, V.~d. Silva, and J.~C. Langford, ``A global geometric
  framework for nonlinear dimensionality reduction,'' \emph{Science}, vol. 290,
  no. 5500, pp. 2319--2323, 2000. [Online]. Available:
  \url{http://science.sciencemag.org/content/290/5500/2319}
\BIBentrySTDinterwordspacing

\bibitem{RoSa00}
S.~Roweis and L.~Saul, ``Nonlinear dimensionality reduction by locally linear
  embedding,'' \emph{Science}, vol. 290, pp. 2323--2326, 2000.

\bibitem{KFKP18}
H.~Kvinge, E.~Farnell, M.~Kirby, and C.~Peterson, ``A gpu-oriented algorithm
  design for secant-based dimensionality reduction,'' in \emph{The 17th IEEE
  International Symposium on Parallel and Distributed Computing}.\hskip 1em
  plus 0.5em minus 0.4em\relax IEEE, 2007, to appear.

\bibitem{bjorck1973numerical}
È.~Bj{\"o}rck and G.~H. Golub, ``Numerical methods for computing angles between
  linear subspaces,'' \emph{Mathematics of computation}, vol.~27, no. 123, pp.
  579--594, 1973.

\bibitem{broomhead_jones_king87}
D.~S. Broomhead, R.~Jones, and G.~P. King, ``Topological dimension and local
  coordinates from time series data,'' \emph{J. Phys. A: Math. Gen}, vol.~20,
  pp. L563--L569, 1987.

\bibitem{AnHuKi02}
M.~Anderle, D.~Hundley, and M.~Kirby, ``The bilipschitz criterion for mapping
  design in data analysis,'' \emph{Intelligent Data Analysis}, vol.~6, no.~1,
  pp. 85--104, 2002.

\bibitem{kirby_hundley1999}
D.~Hundley and M.~Kirby, ``Estimation of topological dimension,'' in
  \emph{Proceedings of the Third SIAM International Conference on Data Mining},
  San Fransico, 2001, pp. 194--202.

\bibitem{costa2006determining}
J.~A. Costa and A.~O. Hero, ``Determining intrinsic dimension and entropy of
  high-dimensional shape spaces,'' in \emph{Statistics and Analysis of
  Shapes}.\hskip 1em plus 0.5em minus 0.4em\relax Springer, 2006, pp. 231--252.

\bibitem{fukunaga1971algorithm}
K.~Fukunaga and D.~R. Olsen, ``An algorithm for finding intrinsic
  dimensionality of data,'' \emph{IEEE Transactions on Computers}, vol. 100,
  no.~2, pp. 176--183, 1971.

\bibitem{BK05}
\BIBentryALTinterwordspacing
D.~S. Broomhead and M.~J. Kirby, ``Dimensionality reduction using secant-based
  projection methods: The induced dynamics in projected systems,''
  \emph{Nonlinear Dynamics}, vol.~41, no.~1, pp. 47--67, Aug 2005. [Online].
  Available: \url{https://doi.org/10.1007/s11071-005-2792-1}
\BIBentrySTDinterwordspacing

\bibitem{camastra2003data}
F.~Camastra, ``Data dimensionality estimation methods: a survey,''
  \emph{Pattern recognition}, vol.~36, no.~12, pp. 2945--2954, 2003.

\bibitem{cunningham2015linear}
J.~P. Cunningham and Z.~Ghahramani, ``Linear dimensionality reduction: Survey,
  insights, and generalizations,'' \emph{The Journal of Machine Learning
  Research}, vol.~16, no.~1, pp. 2859--2900, 2015.

\bibitem{wang2015survey}
F.~Wang and J.~Sun, ``Survey on distance metric learning and dimensionality
  reduction in data mining,'' \emph{Data Mining and Knowledge Discovery},
  vol.~29, no.~2, pp. 534--564, 2015.

\bibitem{Fal03}
\BIBentryALTinterwordspacing
K.~Falconer, \emph{Fractal geometry}, 2nd~ed.\hskip 1em plus 0.5em minus
  0.4em\relax John Wiley \& Sons, Inc., Hoboken, NJ, 2003, mathematical
  foundations and applications. [Online]. Available:
  \url{https://doi.org/10.1002/0470013850}
\BIBentrySTDinterwordspacing

\bibitem{kirby_1998a}
D.~Broomhead and M.~Kirby, ``A new approach for dimensionality reduction:
  Theory and algorithms,'' \emph{SIAM J. of Applied Mathematics}, vol.~60,
  no.~6, pp. 2114--2142, 2000.

\bibitem{H33}
H.~Hotelling, ``Analysis of a complex of statistical variables into principal
  components.'' \emph{Journal of educational psychology}, vol.~24, no.~6, p.
  417, 1933.

\bibitem{J86}
I.~T. Jolliffe, ``Principal component analysis and factor analysis,'' in
  \emph{Principal component analysis}.\hskip 1em plus 0.5em minus 0.4em\relax
  Springer, 1986, pp. 115--128.

\bibitem{HB06}
R.~J. Hathaway and J.~C. Bezdek, ``Extending fuzzy and probabilistic clustering
  to very large data sets,'' \emph{Computational Statistics \& Data Analysis},
  vol.~51, no.~1, pp. 215--234, 2006.

\bibitem{CTZ06}
F.~Cao, A.~K. Tung, and A.~Zhou, ``Scalable clustering using graphics
  processors,'' in \emph{International Conference on Web-Age Information
  Management}.\hskip 1em plus 0.5em minus 0.4em\relax Springer, 2006, pp.
  372--384.

\bibitem{WZH09}
R.~Wu, B.~Zhang, and M.~Hsu, ``Clustering billions of data points using
  {GPU}s,'' in \emph{Proceedings of the combined workshops on UnConventional
  high performance computing workshop plus memory access workshop}.\hskip 1em
  plus 0.5em minus 0.4em\relax ACM, 2009, pp. 1--6.

\bibitem{LZCL13}
Y.~Li, K.~Zhao, X.~Chu, and J.~Liu, ``Speeding up k-means algorithm by
  {GPU}s,'' \emph{Journal of Computer and System Sciences}, vol.~79, no.~2, pp.
  216--229, 2013.

\bibitem{HLDZH09}
B.~Hong-Tao, H.~Li-li, O.~Dan-tong, L.~Zhan-shan, and L.~He, ``K-means on
  commodity gpus with cuda,'' in \emph{Computer Science and Information
  Engineering, 2009 WRI World Congress on}, vol.~3.\hskip 1em plus 0.5em minus
  0.4em\relax IEEE, 2009, pp. 651--655.

\bibitem{SDT08}
S.~A. Shalom, M.~Dash, and M.~Tue, ``Efficient k-means clustering using
  accelerated graphics processors,'' in \emph{International conference on data
  warehousing and knowledge discovery}.\hskip 1em plus 0.5em minus 0.4em\relax
  Springer, 2008, pp. 166--175.

\bibitem{FRCC08}
R.~Farivar, D.~Rebolledo, E.~Chan, and R.~H. Campbell, ``A parallel
  implementation of k-means clustering on gpus.'' in \emph{Pdpta}, vol.~13,
  no.~2, 2008, pp. 212--312.

\bibitem{SAWH14}
A.~S. Shirkhorshidi, S.~Aghabozorgi, T.~Y. Wah, and T.~Herawan, ``Big data
  clustering: a review,'' in \emph{International Conference on Computational
  Science and Its Applications}.\hskip 1em plus 0.5em minus 0.4em\relax
  Springer, 2014, pp. 707--720.

\bibitem{FATAKZFB14}
A.~Fahad, N.~Alshatri, Z.~Tari, A.~Alamri, I.~Khalil, A.~Y. Zomaya, S.~Foufou,
  and A.~Bouras, ``A survey of clustering algorithms for big data: Taxonomy and
  empirical analysis,'' \emph{IEEE transactions on emerging topics in
  computing}, vol.~2, no.~3, pp. 267--279, 2014.

\bibitem{SS03}
H.~Schwetlick and U.~Schnabel, ``Iterative computation of the smallest singular
  value and the corresponding singular vectors of a matrix,'' \emph{Linear
  algebra and its applications}, vol. 371, pp. 1--30, 2003.

\bibitem{H01}
G.~Hongbin, ``Irr: An algorithm for computing the smallest singular value of
  large scale matrices,'' \emph{International Journal of Computer Mathematics},
  vol.~77, no.~1, pp. 89--104, 2001.

\bibitem{LC15}
N.~Lee and A.~Cichocki, ``Estimating a few extreme singular values and vectors
  for large-scale matrices in tensor train format,'' \emph{SIAM Journal on
  Matrix Analysis and Applications}, vol.~36, no.~3, pp. 994--1014, 2015.

\bibitem{Matlab}
``Matlab and statistics toolbox release,'' 2018, the MathWorks, Natick, MA,
  USA.

\bibitem{BR05}
J.~Baglama and L.~Reichel, ``Augmented implicitly restarted lanczos
  bidiagonalization methods,'' \emph{SIAM Journal on Scientific Computing},
  vol.~27, no.~1, pp. 19--42, 2005.

\bibitem{L98}
R.~M. Larsen, ``Lanczos bidiagonalization with partial reorthogonalization,''
  \emph{DAIMI Report Series}, vol.~27, no. 537, 1998.

\bibitem{IP}
{Grupo de Inteligencia Computacional}, ``Hyperspectral remote sensing scenes,''
  2014,
  \url{http://www.ehu.eus/ccwintco/index.php/Hyperspectral_Remote_Sensing_Scenes},
  Last accessed on 2018-4-30.

\end{thebibliography}

\end{document}